\documentclass[11pt,a4paper]{article}
\usepackage[T1]{fontenc}
\usepackage[hyperref]{emnlp2020}
\usepackage{times}
\usepackage{latexsym}
\usepackage{xspace}
\usepackage{graphicx}

\usepackage{gb4e} 
\noautomath
\usepackage{listings}
\DeclareFixedFont{\ttb}{T1}{txtt}{bx}{n}{12} 
\DeclareFixedFont{\ttm}{T1}{txtt}{m}{n}{12}  

\usepackage{amssymb}
\usepackage{pifont} 
\newcommand{\xmark}{\ding{55}}%
\usepackage{pythonhighlight}
\usepackage{float}
\usepackage{subcaption}
\makeatletter
\newcommand\nocaption{%
    \renewcommand\p@subfigure{}
    \renewcommand\thesubfigure{\thefigure\alph{subfigure}}
}
\makeatother

\usepackage{enumitem}

\usepackage{color}
\definecolor{deepblue}{rgb}{0,0,0.5}
\definecolor{deepred}{rgb}{0.6,0,0}
\definecolor{deepgreen}{rgb}{0,0.5,0}

\usepackage{microtype}

\aclfinalcopy 

\newcommand\diagnnose{\mbox{\texttt{diagNNose}}\xspace}
\newcommand\ttbold{\tt\fontseries{b}\selectfont}

\title{\diagnnose: A Library for Neural Activation Analysis}

\author{Jaap Jumelet \\
  Institute for Logic, Language and Computation, University of Amsterdam \\
  \texttt{j.w.d.jumelet@uva.nl} \\
  }

\date{}

\begin{document}
\maketitle
\begin{abstract}
In this paper we introduce \diagnnose, an open source library for analysing the activations of deep neural networks.
\diagnnose contains a wide array of interpretability techniques that provide fundamental insights into the inner workings of neural networks.
We demonstrate the functionality of \diagnnose with a case study on subject-verb agreement within language models.
\diagnnose is available at \url{https://github.com/i-machine-think/diagnnose}.
\end{abstract}

\section{Introduction}
We introduce \diagnnose, an open source library for analysing deep neural networks.
The \diagnnose library allows researchers to gain better insights into the internal representations of such networks, providing a broad set of tools of state-of-the-art analysis techniques.
The library supports a wide range of model types, with a main focus on NLP architectures based on LSTMs \citep{hochreiter1997long} and Transformers \citep{vaswani2017attention}.

Open-source libraries have been quintessential in the progress and democratisation of NLP.
Popular packages include
HuggingFace's \texttt{transformers} \citep{Wolf2019HuggingFacesTS} -- allowing easy access to pre-trained Transformer models;
\texttt{jiant} \citep{pruksachatkun2020jiant} -- focusing on multitask and transfer learning within NLP;
\texttt{Captum} \citep{captum} -- providing a range of feature attribution methods;
and \texttt{LIT} \citep{lit} -- a platform for visualising and understanding model behaviour.
We contribute to the open-source community by incorporating several \mbox{\textbf{interpretability}} techniques that have not been present in these packages.

Recent years have seen a considerable interest in improving the understanding of how deep neural networks operate \citep{linzen2019proceedings}.
The high-dimensional nature of these models makes it notoriously challenging to untangle their inner dynamics.
This has given rise to a novel subfield within AI that focuses on interpretability, providing us a peak inside the black box.
\diagnnose aims to unify several of these techniques into one library, allowing interpretability research to be conducted in a more streamlined and accessible manner.

\diagnnose's main focus lies on techniques that aid in uncovering linguistic knowledge that is encoded within a model's representations.
The library provides abstractions that allow recurrent models to be investigated in the same way as Transformer models, in a modular fashion.
It contains an extensive \textbf{activation extraction} module that allows for the extraction of (intermediate) model activations on a corpus.
The analysis techniques that are currently implemented include: \begin{itemize}[leftmargin=.5cm]
  \setlength\itemsep{0em}
    \item \textbf{Targeted syntactic evaluation tasks}, such as those of \citet{linzen-etal-2016-assessing} and \citet{marvin-linzen-2018-targeted}.
    \item \textbf{Probing} with \textbf{diagnostic classifiers} \citep{hupkes2018visualisation, adi2016fine}, and \textbf{control tasks} \citep{hewitt2019designing}.
    \item \textbf{Feature attributions} that retrieve a feature's contribution to a model prediction \citep{lundberg2017unified, murdoch18cd}. 
    Our implementation is model-agnostic, which means that any type of model architecture can be explained by it.

\end{itemize}


In this paper we present both an overview of the library, as well as a case study on subject-verb agreement within language models.
We first present a brief overview of interpretability within NLP and a background to the analysis techniques that are part of the library (Section \ref{sec:background}).
We then provide an overview of \diagnnose and expand briefly on its individual modules (Section \ref{sec:overview}).
We conclude with a case study on subject-verb agreement, demonstrating several of \diagnnose's features in an experimental setup (Section \ref{sec:sva}).

\section{Background} \label{sec:background}
The increasing capacities of language models (and deep learning in general) have led to a rich field of research that aims to gain a better understanding of how these models operate.
Approaches in this research area are often interdisciplinary in nature, borrowing concepts from fields such as psycho-linguistics, information theory, and game theory.
\diagnnose provides support for several influential analysis techniques, for which we provide a brief background here.

\subsection{Targeted syntactic evaluations}
Language models have stood at the basis of many successes within NLP in recent years \citep{peters2018deep, devlin2019bert}.
These models are trained on the objective of predicting the probability of an upcoming (or masked) token.
In order to succeed in this task, these models need to possess a notion of many different linguistic aspects, such as syntax, semantics, and general domain knowledge.
One popular line of research that tries to uncover a model's linguistic capacities does this via so-called {targeted syntactic evaluations} \citep{linzen-etal-2016-assessing, gulordava2018colorless, marvin-linzen-2018-targeted, jumelet2018language}.
This type of analysis compares a model's output on minimally different pairs of grammatical and ungrammatical constructions.
If it assigns a higher probability to the grammatical construction, the model is said to possess a notion of the underlying linguistic principles, such as subject-verb agreement or NPI licensing:
\begin{exe}
    \ex\begin{xlist}
    \ex[]{The \textbf{ladies} near \underline{John} \textbf{walk}.}
    \ex[*]{The \textbf{ladies} near \underline{John} \underline{walks}.}
    \end{xlist}
    
    \ex\begin{xlist}
    \ex[]{\textbf{Nobody} has \textbf{ever} been there.}
    \ex[*]{Someone has \textbf{ever} been there.}
    \end{xlist}
\end{exe}
\diagnnose supports a wide range of syntactic tasks, as well as an interface that allows new tasks to be added without effort.

\subsection{Diagnostic Classifiers}
A second line of work tries to assess a model's understanding of linguistic properties -- such as part-of-speech tags or number information -- by directly training \textbf{diagnostic classifiers} on top of its representations \citep{hupkes2018visualisation, adi2016fine, belinkov2017neural}.
This type of analysis, also referred to as \textbf{probing}, has led to numerous insights into the inner workings of language models \citep{liu2019linguistic, tenney2019bert}.
The activations diagnostic classifiers are trained on are not restricted to just the hidden states of a language model at their top layer: this can, for instance, also be done on the individual gate activations to reveal patterns at the cell-level of a model \citep{giulianelli2018under, lakretz-etal-2019-emergence}.

Recently, it has been a topic of discussion to what extent a high accuracy of a diagnostic classifier signifies that that property is actively being encoded by the model.
Several solutions to assess this have been proposed, such as training a diagnostic classifier on a baseline of random labels (called a \textit{control task} \citep{hewitt2019designing}), or based on the minimum description length of the classifier, a concept from information theory \citep{voita2020information, pimentel2020information}.
\diagnnose currently facilitates the training of diagnostic classifiers, as well as training control tasks alongside them.

\subsection{Feature Attributions}
Although probing allows us to uncover specific properties that are embedded within the model representations, it is unable to explain \textit{how} a model transforms its input features into a successful prediction.
This question can be addressed by computing the input \textbf{feature contributions} to a subsequent output.
This is a challenging task, as the high-dimensional, non-linear nature of deep learning models prevents us from expressing these contributions directly on the basis of the model parameters.

Feature attributions can be computed in different ways.
One common approach to this task is based on a concept that stems from cooperative game theory, called the Shapley value \citep{shapley1953value}.
A Shapley value expresses the contribution of a player (in our case an input feature) to the outcome of game (in our the case a model prediction).
Computing Shapley values is computationally expensive, and several approximation algorithms have therefore been proposed, such as SHAP \citep{lundberg2017unified}, and Integrated Gradients \citep{DBLP:conf/icml/SundararajanTY17}.
\diagnnose currently facilitates the computation of feature attributions using a technique called Contextual Decomposition \citep{murdoch18cd}, and its generalisation as proposed by \citet{jumelet-etal-2019-analysing}.

\section{Library Overview}\label{sec:overview}

\subsection{Modules} 
\begin{figure}
    \centering
    \includegraphics[width=\columnwidth,trim={3cm 3cm 3cm 3cm},clip]{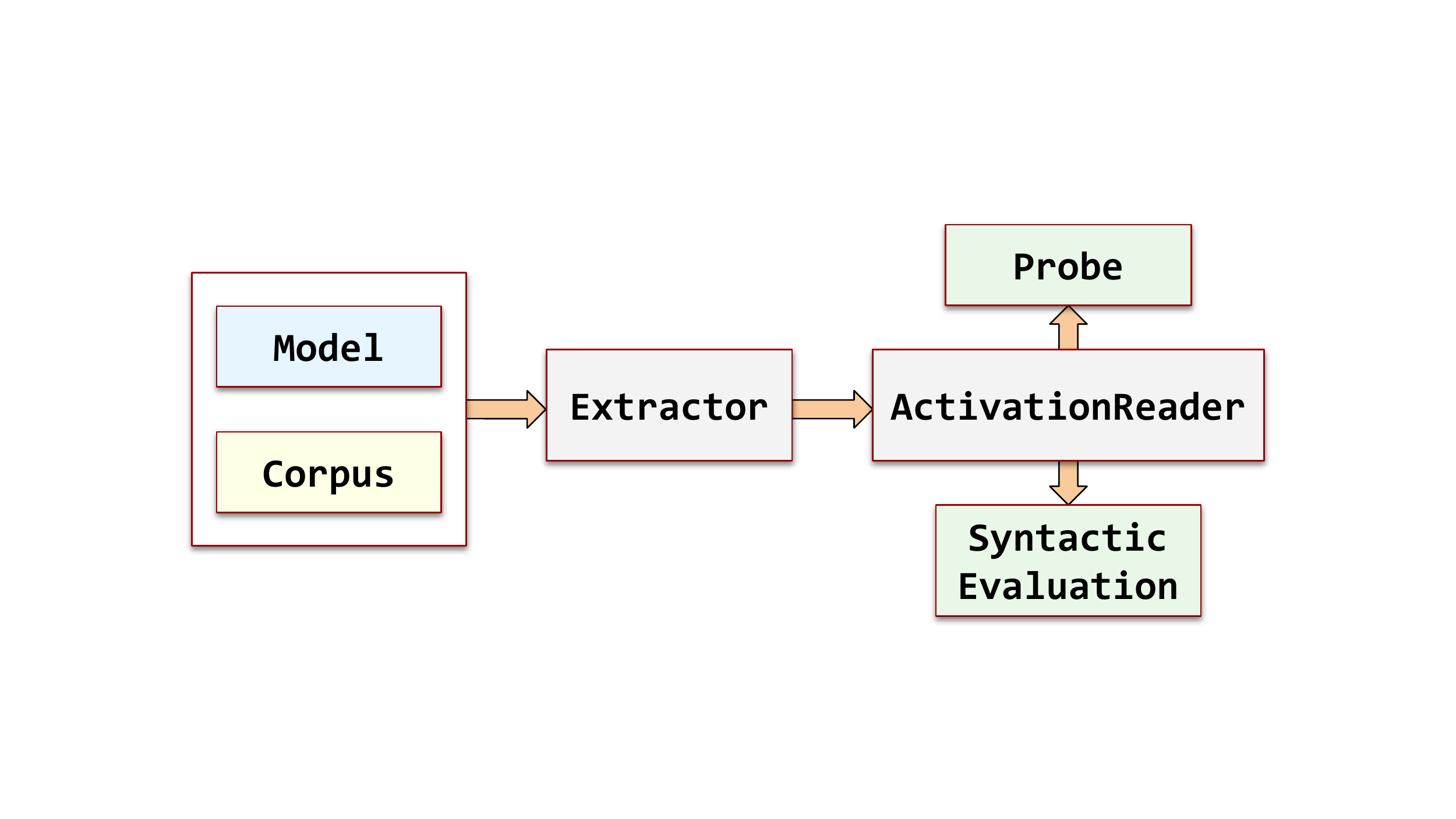}
    \caption{
    Pipeline stages for conducting syntactic evaluation and probing experiments.
    Note the modular nature of the pipeline: activations need only to be extracted once, after which the setup of the analysis experiments can be fine-tuned effortlessly.
    }
    \label{fig:pipeline}
\end{figure}

The library is structured into several modules that can be used as building blocks for an experimental pipeline.
We provide an overview of a possible experimental pipeline in Figure \ref{fig:pipeline}.

\subsubsection{Core modules}
The following core modules stand at the basis of the different pipelines that can be build on top of \diagnnose.

\paragraph{\ttbold models} 
We provide an abstraction over language models, enabling recurrent and Transformer models to derive from the same interface.
Importing pre-trained Transformer models is done via the \texttt{transformers} library.
For recurrent models we provide a wrapper that enables access to intermediate activations, including gate activations.
We also provide functionality that allows to set the initial hidden states of recurrent LMs, based on a sentence or corpus.\footnote{As has been noted by \citet{jumelet-etal-2019-analysing}, LSTM LMs perform better when initialised with the phrase \mbox{``\texttt{. <eos>}''}, instead of zero-valued vectors.}

\paragraph{\ttbold corpus}
Corpora are imported as a \texttt{Dataset} from the \texttt{torchtext} package.
A Corpus can be transformed into an iterator for processing.
Tokenization is done using the \texttt{transformers} tokenizers, allowing tokenization to be done in both a traditional token-per-token fashion, or based on sub-word units, such as byte pair encodings \citep{sennrich2016neural}.

\paragraph{\ttbold extract} 
Central to most of the analysis modules is the extraction of activations.
We provide an \texttt{Extractor} class that can extract the activations of a model given a corpus.
Thanks to our model wrappers activation extraction is not restricted to just the top layer of a model; intermediate (gate) activations can be extracted as well.
To facilitate the extraction of larger corpora with limited computational resources, activations can be dumped dynamically to disk.

\paragraph{\ttbold activations}
Extracted activations can easily be retrieved using a \texttt{ActivationReader}, providing access to activations that correspond to a specific subset of corpus sentences.
We also provide functionality for extracting only a specific subset of activations, based on sentence and token information.
This way it is possible, for instance, to only extract the activations at the position of tokens of particular interest.

\paragraph{\ttbold config}
The pipeline of \diagnnose is configuration-driven.
Configuration is defined in JSON format, but individual attributes can also be set from the command line directly.

\subsubsection{Analysis modules}
\begin{figure*}
    \centering
    \includegraphics[width=\textwidth,trim={0cm 2cm 0cm 2cm},clip]{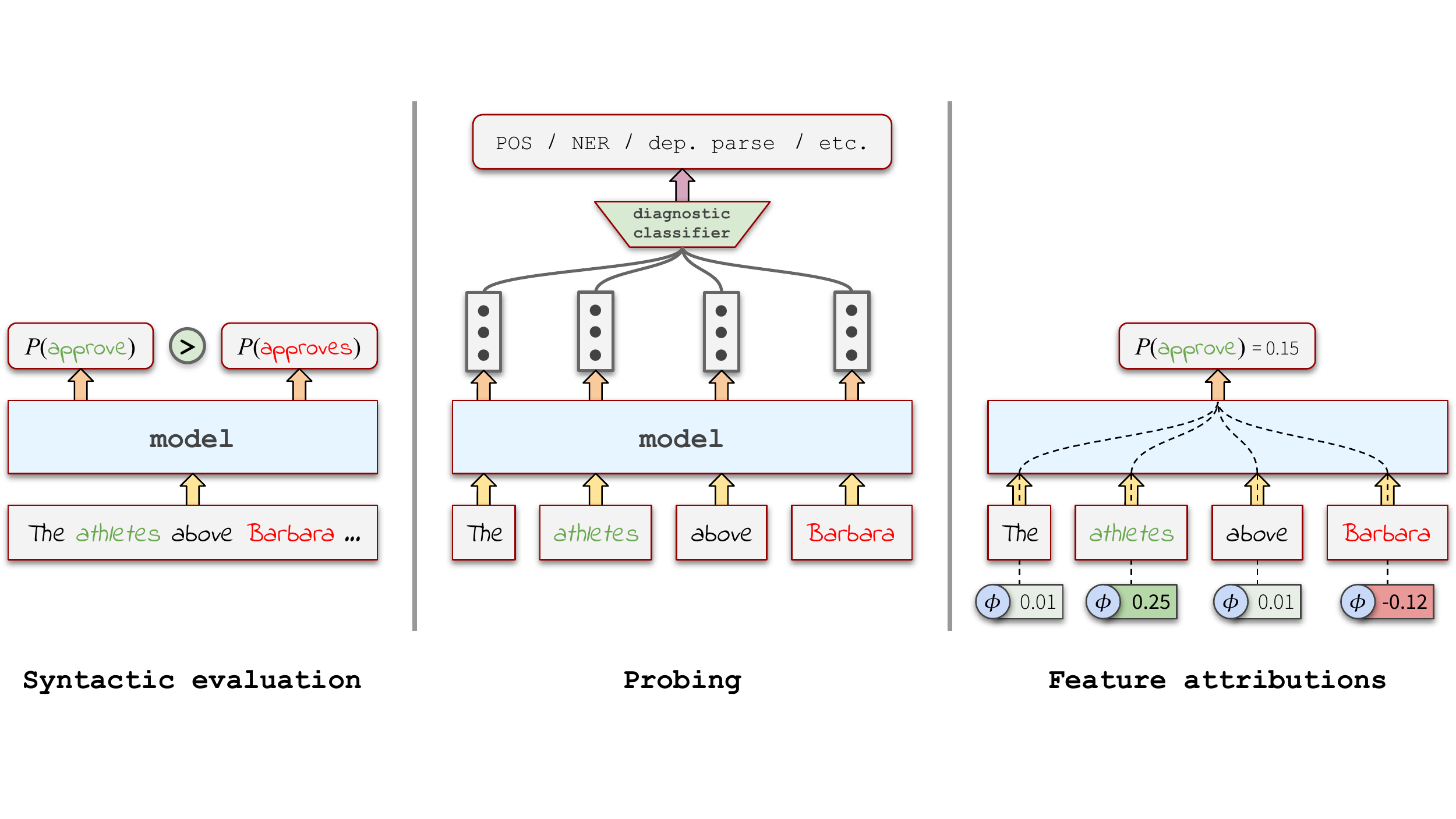}
    \caption{
    Schematic overview of three different types of experiments that are supported by \diagnnose.
    }
    \label{fig:experiments}
\end{figure*}
We currently support three main types of experimental modules.
We provide a graphical overview of these modules in Figure~\ref{fig:experiments}.

\paragraph{\ttbold syntax}
The library provides functionality for a large suite of targeted syntactic evaluation tasks.
Currently we provide support for the following tasks: \begin{itemize}[leftmargin=.5cm]
    \setlength\itemsep{0em}
    \item The subject-verb agreement corpus of \citet{linzen-etal-2016-assessing}, for which we also provide more fine-grained attractor conditions; 
    \item The wide range of linguistic expressions of \citet{marvin-linzen-2018-targeted}; 
    \item The subject-verb agreement tasks of \citet{lakretz-etal-2019-emergence}; 
    \item The NPI corpus of  \citet{warstadt-etal-2019-investigating}; 
    \item The stereotypically gendered anaphora resolution corpus of \citet{jumelet-etal-2019-analysing}, based on the original WinoBias corpus of \citet{zhao2018gender}.
\end{itemize}
Furthermore, the current implementation permits similar types of tasks to be easily added, and we plan on incorporating a larger set of tasks in the near future.

\paragraph{\ttbold probe}
We provide easy tooling for training diagnostic classifiers \citep{hupkes2018visualisation, adi2016fine} on top of extracted activations, to probe for linguistic information that might be embedded within them.
Our extraction module facilitates training diagnostic classifiers on top of intermediate activations as well, including gate activations. 
In recent years it has been pointed out that a high probing accuracy does not necessarily imply that linguistic information is actively being encoded by a model.
To address this we have incorporated functionality for Control Tasks \citep{hewitt2019designing}, providing more qualitative insights.

\paragraph{\ttbold attribute}
We provide functionality for model-agnostic feature attributions, that allow the output of a model to be decomposed into a sum of contributions.
This is achieved by implementing a wrapper over the operations of PyTorch\footnote{The wrapper is defined based on the \texttt{\_\_torch\_function\_\_} functionality that has been introduced in PyTorch 1.5.}, allowing intermediate feature contributions to be propagated during a forward pass in the model.
Our implementation provides a basis for many Shapley-based attribution methods, as it allows different approximation methods to be tested easily.
We currently facilitate the approximation procedure of (Generalised) Contextual Decomposition \citep{murdoch18cd, jumelet-etal-2019-analysing} and Shapley sampling values \citep{CASTRO20091726}, next to the exact computation of propagated Shapley values.
Our implementation is the first model-agnostic implementation of Contextual Decomposition: previous implementations were dependent on a fixed model structure.


\subsection{Requirements}
\diagnnose is released on \texttt{pip} and can be installed using \texttt{pip install diagnnose}, or directly cloned from the GitHub repository: \url{https://github.com/i-machine-think/diagnnose}.
The library supports Python 3.6 or later, and its core dependencies are PyTorch \citep{paszke2019pytorch} (v1.5+), \texttt{torchtext}\footnote{\url{https://pytorch.org/text/}}, and HuggingFace's \texttt{transformers} \citep{Wolf2019HuggingFacesTS}.
\diagnnose is released under the MIT License \citep{mit2020license}.
\diagnnose runs both on CPUs and GPUs, and has especially been optimised for smaller consumer setups, due to limited computational resources during development.

The \diagnnose code base is fully typed using Python type hints.
The code is formatted using \textit{Black}.\footnote{\url{https://github.com/psf/black}}
All methods and classes are documented, and an overview of this documentation can be found on \url{https://diagnnose.readthedocs.io}.


\section{Case Study: Subject-Verb Agreement}\label{sec:sva}
To demonstrate the functionality of \diagnnose we will consider the subject-verb agreement corpora of \citet{lakretz-etal-2019-emergence} on a set of language models.
For our experiments we consider the following models: BERT \citep{devlin2019bert}, RoBERTa \citep{liu2019roberta}, DistilRoBERTa \citep{sanh2019distilbert}, and the LSTM language model of \citet{gulordava2018colorless}.

\subsection{Corpora}
The corpora of \citet{lakretz-etal-2019-emergence} are formed by seven tasks of template-based syntactic constructions.
These constructions contain an ``agreement attractor'' in between the subject and the verb, which might trick a language model into predicting the incorrect number of the verb.
A model thus needs to possess a strong notion of the structure of a sentence: nouns within a prepositional phrase, for instance, should have no impact on the number of the main verb in a sentence.

The seven tasks are defined by the following templates:\begin{itemize}[leftmargin=0cm]
    \setlength\itemsep{0em}
    \item[] \textsc{Simple}\\ The \textbf{athletes} \textbf{approve} 
    \item[] \textsc{Adv}\\ The \textbf{uncle} probably \textbf{avoids} 
    \item[] \textsc{2Adv}\\ The \textbf{athlete} most probably \textbf{understands} 
    \item[] \textsc{CoAdv}\\ The \textbf{farmer} overtly and deliberately \textbf{knows} 
    \item[] \textsc{NamePP}\\ The \textbf{women} near John \textbf{remember} 
    \item[] \textsc{NounPP}\\ The \textbf{athlete} beside the tables \textbf{approves} 
    \item[] \textsc{NounPPAdv}\\ The \textbf{aunt} behind the bikes certainly \textbf{knows} 
\end{itemize}

Each task contains 600 to 900 distinct sentences.
Sentences are split up into multiple conditions based on the number of the subject, and the number of the intervening noun phrase.
The \textsc{NounPP} corpus, for instance, is split up into 4 conditions: \begin{itemize}
    \setlength\itemsep{0em}
    \item[SS:] The \textbf{athlete} beside the table \textbf{approves} 
    \item[SP:] The \textbf{athlete} beside the tables \textbf{approves} 
    \item[PS:] The \textbf{athletes} beside the table \textbf{approves} 
    \item[PP:] The \textbf{athletes} beside the tables \textbf{approves} 
\end{itemize}

\begin{figure}
    \nocaption
    \centering
    \begin{subfigure}[b]{\columnwidth}
    \includegraphics[width=\columnwidth]{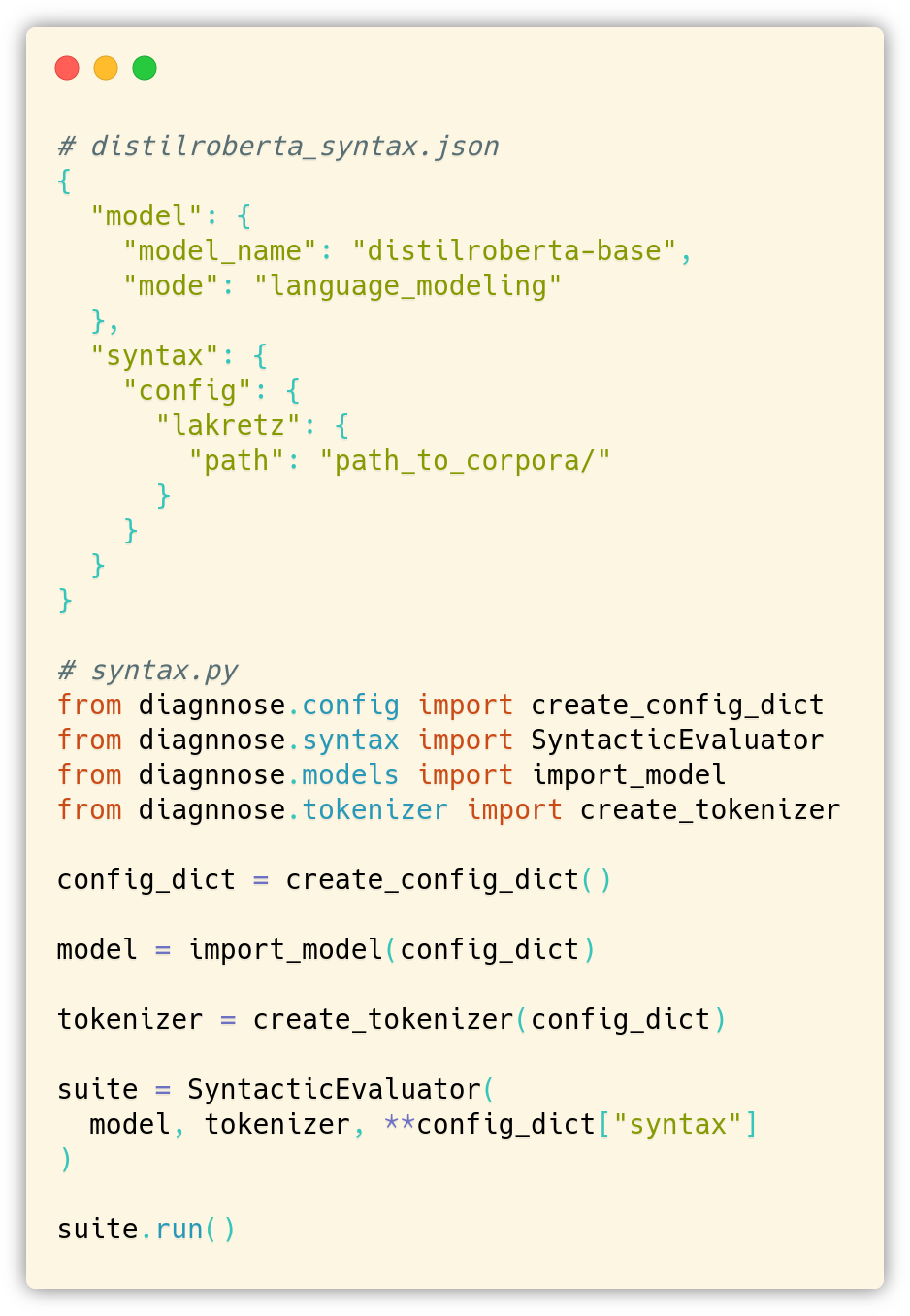}
    \caption{Example setup for running the targeted syntactic evaluation tasks of \citet{lakretz-etal-2019-emergence} on DistilRoBERTa.}\label{fig:syntax}
    \end{subfigure}\\[10pt]

    \begin{subfigure}[b]{\columnwidth}
    \includegraphics[width=\columnwidth]{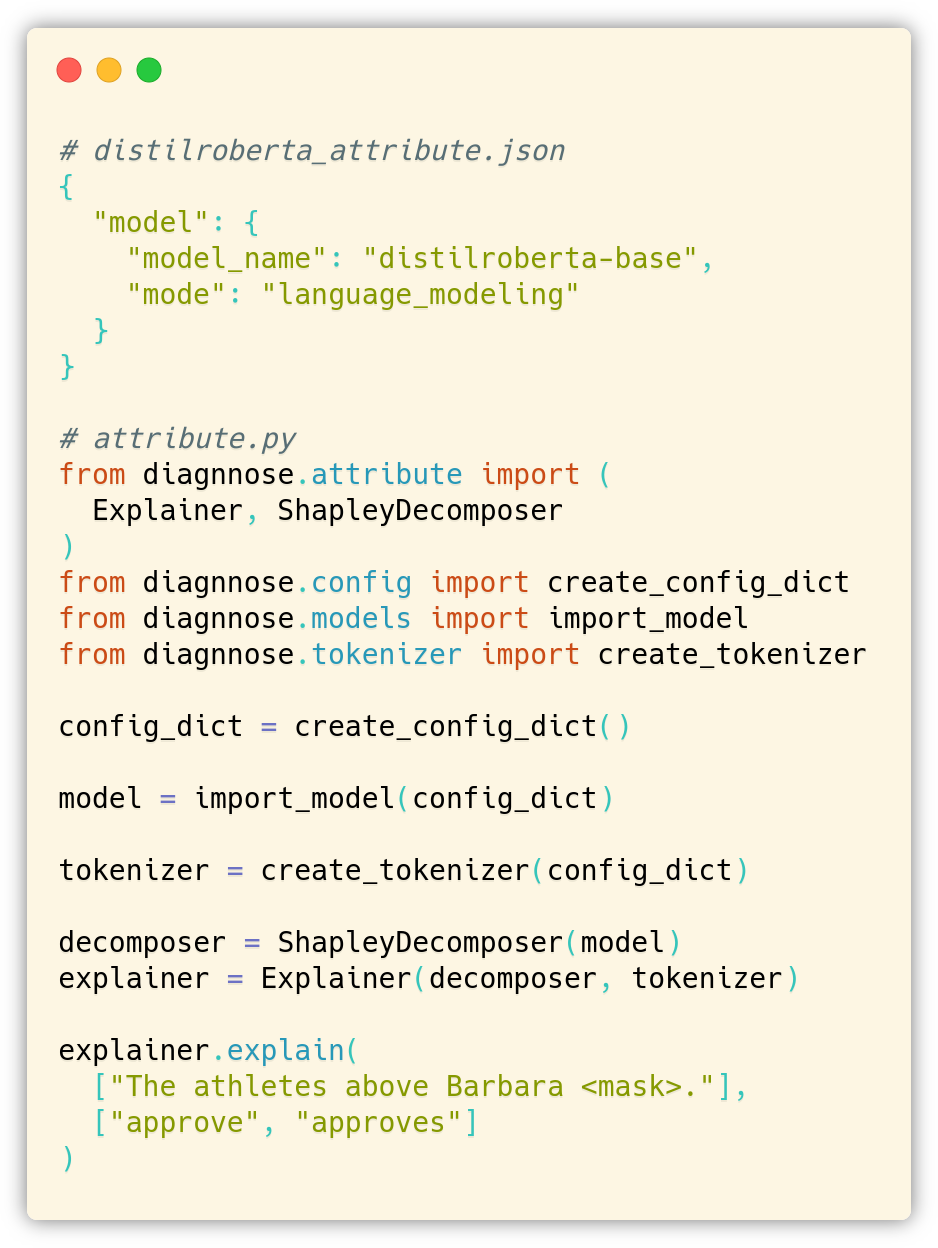}
    \caption{Example setup for creating the feature attributions of DistilRoBERTa on a sentence from the \textsc{NamePP} corpus of \citet{lakretz-etal-2019-emergence}.}\label{fig:attribute}
    \end{subfigure}
    \label{fig:code}
\end{figure}

\begin{table*}[ht!]
    \centering
    \begin{tabular}{|l|c||c|c|c|c|}
        \hline
        Corpus & Condition & BERT & RoBERTa & DistilRoBERTa & LSTM \\\hline
        \textsc{Simple} & S & \textbf{100} & \textbf{100} & \textbf{100} & \textbf{100}\\
        & P & \textbf{100} & \textbf{100} & \textbf{100} & \textbf{100}\\\hline
        \textsc{Adv} & S & \textbf{100} & \textbf{100} & \textbf{100} & \textbf{100}\\
        & P & \textbf{100} & \textbf{100} & \textbf{100} & 99.6\\\hline
        \textsc{2Adv} & S & \textbf{100} & \textbf{100} & \textbf{100} & 99.2\\ 
        & P & \textbf{100} & \textbf{100} & \textbf{100} & 99.3\\\hline
        \textsc{CoAdv} & S & \textbf{100} & \textbf{100} & \textbf{100} & 98.7\\
        & P & \textbf{100} & \textbf{100} & \textbf{100} & 99.3\\\hline
        \textsc{NamePP} & SS & 93.0 & 75.7 & 81.5 & \textbf{99.3}\\
        & PS & \textbf{88.4} & 65.9 & 32.4 & 68.9\\\hline
        \textsc{NounPP} & SS & 95.7 & 88.9 & 98.1 & \textbf{99.2}\\
        & SP & \textbf{93.3} & 84.7 & 91.1 & 87.2\\
        & PS & \textbf{96.7} & 90.6 & 85.3 & 92.0\\
        & PP & \textbf{100} & \textbf{100} & \textbf{100} & 99.0 \\\hline
        \textsc{NounPPAdv} & SS & 99.6 & \textbf{100} & \textbf{100} & 99.5\\
        & SP & 99.2 & 99.8 & \textbf{100} & 91.2\\
        & PS & \textbf{100} & \textbf{100} & \textbf{100} & 99.2\\
        & PP & \textbf{100} & \textbf{100} & \textbf{100} & 99.8 \\\hline
    \end{tabular}
    \caption{Results of the targeted syntactic evaluation tasks of \citet{lakretz-etal-2019-emergence}.
    }
    \label{tab:syntax}
\end{table*}

To test these corpora on a recurrent model, we first compute the model's hidden state at the position of the verb by feeding it the sub-sentence up till that position.
Based on this hidden state we compute the output probabilities of the verb of the correct number ($v^{\checkmark}$), and the incorrect number ($v^{\textup{\xmark}}$), and compare these:
\[P(v^{\checkmark}~|~h_t) > P(v^{\textup{\xmark}}~|~h_t)\]
For bi-directional masked language models, such as BERT, we can not compute a model's intermediate hidden state by passing it a sub-sentence, because these models also incorporate the input of future tokens.
To solve this, we replace the verb in each sentence with a \texttt{<mask>} token, and assess the model's probabilities at the position of this token.
Modern language models often make use of BPE tokenization that might split a word into multiple sub-words.
In our experiments we therefore only compare verb forms for which both the plural and singular form are split into a single token.\footnote{The RoBERTa tokenizer, for example, splits ``confuses'' into ``conf'' + ``uses'', and ``confuse'' into ``confuse''. Comparing the model probabilities for these two forms directly is hence not possible.}

\subsection{Targeted syntactic evaluations}
We run the targeted syntactic evaluation suite on all the 7 templates.
An example configuration and script of this experiment in provided in Figure \ref{fig:syntax}.
To run the experiment on a different model, the only configuration that needs to be changed is the \texttt{model\_name}.
The results of the experiment are shown in Table \ref{tab:syntax}.

It can be seen that the Transformer language models generally achieve higher scores than the LSTM model. 
Interestingly, the \textsc{NamePP} task poses a challenge for all models, and both RoBERTa and DistilRoBERTa score lower on this task than the LSTM.
A second point of interest is the difference in performance between RoBERTa and DistilRoBERTa on the \textsc{NamePP} and \textsc{NounPP} tasks. 
Even though DistilRoBERTa has been trained to emulate the  behaviour of RoBERTa, its performance on a downstream task like this differs significantly.
These results can provide a starting point for a more fine-grained analysis, such as creating the feature attributions of a model on a specific template.

\subsection{Feature attributions}
\begin{figure*}
    \centering
    \includegraphics[width=\textwidth]{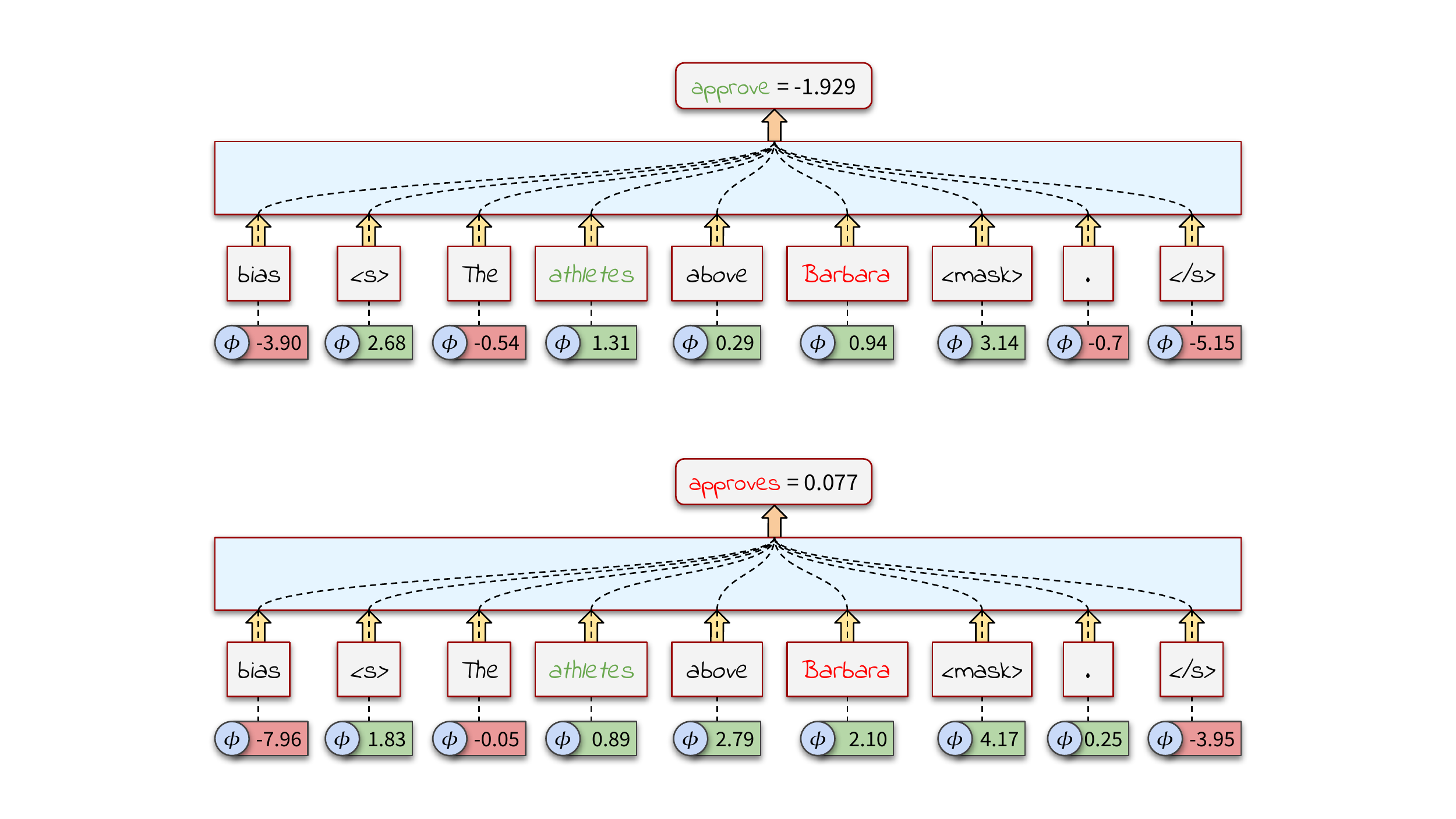}
    \caption{
    The feature attributions for DistilRoBERTa on an example sentence from the \textsc{NamePP} task of \citet{lakretz-etal-2019-emergence}.
    The logits of two output tokens, `\textit{approve}' and `\textit{approves}', are decomposed into a sum of contributions.
    }
    \label{fig:results}
\end{figure*}
To gain a better insight into why the language models struggle so strongly with the \textsc{NamePP} corpus, we run the feature attribution module on these constructions.
An example configuration of this experimental setup is provided in Figure \ref{fig:attribute}.
The results for the experiment are shown in Figure \ref{fig:results}.

We show the attributions for DistilRoBERTa on an example sentence from the corpus, which highlights the difference in impact of the intervening attractor on the number of the verb.
The results should be interpreted as follows: the score at the top of the attribution denotes the \textit{full logit} of the model for that class, these are the logits that are transformed into probabilities using SoftMax.
This logit is \textit{decomposed} in a sum of contributions, which we denote at the bottom of each token.
It can be validated that the contributions sum up together to the logit.
This is an important property of feature attribution methods -- called \textit{efficiency} -- that warrants a certain degree of faithfulness of an explanation to the model.
A negative value indicates a negative feature contribution to an output class: the impact of that feature led to a decreased preference for the class.
Feature attributions also include the influence of model biases: an aggregate of all information that is statically present within the network such as weight intercepts.

On the presented example sentence, DistilRoBERTa makes an incorrect prediction: the logit of the incorrect singular form `\textit{approves}' is larger than that of the plural `\textit{approve}'.
The model's misstep in predicting the correct verb form arrives from the fact that the subject `\textit{athletes}' provided not enough contribution to overrule the negative contributions stemming from other input features.
A model that has a thorough understanding of subject-verb agreement should assign a larger contribution to the subject when predicting the main verb: the number signal provided by the subject should be propagated strongly enough to overrule other interfering signals.

The \texttt{attribute} module is still in active development.
The exponential nature of computing Shapley values makes creating these explanations a challenging task, and we look forward to incorporate other techniques that aim to alleviate the computing costs.

\section{Conclusion}
\diagnnose provides essential tools for conducting interpretability research, providing cutting edge analysis techniques such as diagnostic classifiers and feature attributions.
The modular design of the library allows complex hypotheses to be tested rapidly, and provides a solid basis for the development of novel interpretability techniques.
The library code is open source and welcomes others to contribute: we are eagerly looking forward to collaborate on adding new features to the library.

\section*{Acknowledgments}
The author gratefully acknowledges the feedback received from Dieuwke Hupkes during the development of the library. 

\bibliography{anthology,emnlp2020}
\bibliographystyle{acl_natbib}

\appendix

\end{document}